\def\year{2016}\relax
\documentclass[letterpaper]{article}
\usepackage{aaai16}
\usepackage{times}
\usepackage{helvet}
\usepackage{courier}
\usepackage{color}
\usepackage{graphicx}
\usepackage{multirow}
\usepackage{epsfig}
\usepackage{amsmath}
\usepackage{amssymb}
\usepackage{subfigure}
\usepackage{caption}
\usepackage[font=small,skip=4pt]{caption}
\usepackage{multicol}
\begin{document}
\title{Reading Scene Text in Deep Convolutional Sequences}
\newcommand*\samethanks[1][\value{footnote}]{\footnotemark[#1]}
\author{Pan He\thanks{Authors contributed equally}$^{1,}$ $^2$, Weilin Huang\samethanks$^{1,}$ $^2$, Yu Qiao$^1$, Chen Change Loy$^{2,}$ $^1$, and Xiaoou Tang$^{2,}$ $^1$
       \\
       $^1$Shenzhen Key Lab of Comp. Vis and Pat. Rec.,
       \\
       Shenzhen Institutes of Advanced Technology, Chinese Academy of Sciences, China
       \\
       $^2$Department of Information Engineering, The Chinese University of Hong Kong\\
       \texttt{\{pan.he,wl.huang,yu.qiao\}@siat.ac.cn, \{ccloy,xtang\}@ie.cuhk.edu.hk}
}

\maketitle
\begin{abstract}
\begin{quote}
We develop a Deep-Text Recurrent Network (DTRN) that regards scene text reading as a sequence labelling problem. We leverage recent advances of deep convolutional neural networks to generate an ordered high-level sequence from a whole word image, avoiding the difficult character segmentation problem.  Then a deep recurrent model, building on long short-term memory (LSTM), is developed to robustly recognize the generated CNN sequences, departing from most existing approaches recognising each character independently. Our model has a number of appealing properties in comparison to existing scene text recognition methods: (i) It can recognise highly ambiguous words by leveraging meaningful context information, allowing it to work reliably without either pre- or post-processing; (ii) the deep CNN feature is robust to various image distortions; (iii) it retains the explicit order information in word image, which is essential to discriminate word strings; (iv) the model does not depend on pre-defined dictionary, and it can process unknown words and arbitrary strings.
It achieves impressive results on several benchmarks, advancing  the-state-of-the-art substantially.

\end{quote}
\end{abstract}

\begin{figure}
\begin{center}
\includegraphics[height=7.5cm,width=8cm]{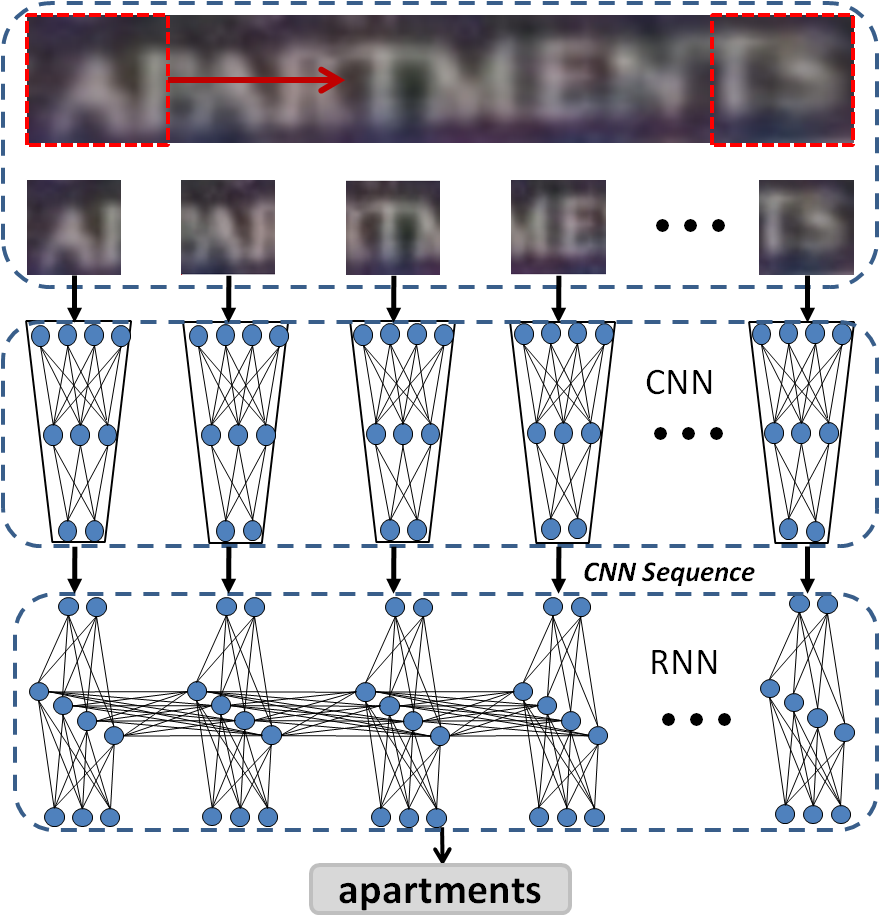}
\end{center}
%\vskip -0.2cm
   \caption{The word image recognition pipeline of the proposed\emph{ Deep-Text Recurrent Networks (DTRN)} model.}
\label{fig:pipeline}
\end{figure}

%With the increasing popularity of cameras-embedded mobile devices and the development of internet sharing capabilities, it has been more challenge for indexing and extracting informative content from a huge amount of multimedia data.
 Text recognition in natural image has received increasing attention in computer vision and machine intelligence, due to its numerous practical applications. This problem includes two sub tasks, namely text detection \cite{Huang2014,Yin2014,He2015,Zhang2015,Huang2013,Neumann2013} and text-line/word recognition \cite{Jaderberg2014,Almazan2014,Jaderberg2014b,Bissacco2013,Yao2014}. This work focuses on the latter that aims to retrieve a text string from a cropped word image.
Though huge efforts have been devoted to this task, reading text in unconstrained environment is still extremely challenging, and remains an open problem, as substantiated in recent
literature~\cite{Jaderberg2015b,Almazan2014}.
%literature~\cite{Jaderberg2015b,Almazan2014,Yao2014,Jaderberg2014,Bissacco2013}.
%
The main difficulty arises from the large diversity of text patterns (e.g. low resolution, low contrast, and blurring), and highly complicated background clutters.
Consequently, individual character segmentation or separation is extremely challenging.

% The segmentation problem sets up the main gap between current word image reading and traditional documented recognition where text information can be easily segmented or binaried, and well-developed OCR technologies have been adopted.

%such as strong lighting, large-scale occlusion and a large amount of noise or outliers
Most previous studies focus on developing powerful character classifiers, some of which are incorporated with a language model, leading to the state-of-the-art performance \cite{Jaderberg2014,Bissacco2013,Yao2014,Lee2014}.
These approaches mainly follow the pipeline of conventional OCR techniques by first involving a character-level segmentation, then followed by an isolated character classifier and post-processing for recognition.
They also adopt deep neural networks for representation learning, but the recognition is still confined to character-level classification.
%
%For example, Bissacco \emph{et al.}\cite{Bissacco2013} proposed a PhotoOCR system by designing a five-layer neural network for character recognition.
%
%In \cite{Jaderberg2014}, a deep CNN was trained for multiple tasks by sharing features.
%
%Note that all these current successful scene text recognition systems are mostly built on isolated character classifier.
Thus their performance are severely harmed by the difficulty of character segmentation or separation. Importantly, recognizing each character independently discards meaningful context information of the words, significantly reducing its reliability and robustness.

First, we wish to address the issue of context information learning.
The main inspiration for approaching this issue comes from the recent success of recurrent neural networks (RNN) for handwriting recognition \cite{Graves2009,Graves2008}, speech recognition \cite{Graves2014}, and language translation \cite{Sutskever2014}.
We found the strong capability of RNN in learning continuous sequential features particularly well-suited for text recognition task to retain the meaningful interdependencies of the continuous text sequence.
We note that RNNs have been formulated for recognizing handwritten or documented images~\cite{Graves2009,Graves2008,Breuel2013}, nevertheless, the background in these tasks is relatively plain, and the raw image feature can be directly input to RNN for recognition, or the text stroke information can be easily extracted or binarized at pixel level, making it possible to manually design a sequential heuristic feature for the input to RNN.
In contrast, the scene text image is much more complicated where pixel-level segmentation is extremely difficult, especially for highly ambiguous images (Fig.~\ref{fig:pipeline}). Thus it is non-trivial to directly apply the sequence labelling models to scene text.

% Though a number of approaches has been developed to segment the text pixels from background, their performance are still far from those working on handwritten or documented images \cite{}.

Consequently, the second challenge we need to resolve is the issue of character segmentation.
We argue that individual character segmentation is not a `must' in text recognition. The key is to acquire strong representation from the image, with explicit order information. The strong representation ensures robustness to various distortions and background clutters, whilst the explicit order information is crucial to discriminate a meaningful word.
%The ordered strong feature sequence can then be read by any sequence labeling model for recognition, avoiding the character segmentation.
The ordered strong feature sequence computed from the sequential regions of word image allows each frame region to locate the part of a character, which can be stored sequentially by the recurrent model. This makes it possible to recognize the character robustly by using its continuous parts, and thus successfully avoid the character segmentation.

% Our work is also motivated from recent success of the RNN models for image caption \cite{Andrej2015,Donahue2014,Alsharif2013}. These systems apply the CNN model for computing a high-level deep feature from a whole image, followed by a RNN model to decode the deep feature into a sequence of words (a sentence) recurrently. But our task is different from them. The word images include explicit order information of its string, which is a crucial cue to discriminate a word.  For the image caption, the image and its target captions do not have such strictly spatial correlation. Global representation may include implicit spatial information of the objects in the image, but applying it to the word image would significantly loss the strict order information. Our goal here is to derive a set of robust sequential features from the word image, and design an new model that successfully connects image representation to the sequence labelling task.

% Therefore, representing the whole word image with a single high-level feature may loss the explicit order information, leading to the loss in discrimination.

To this end, we develop a deep recurrent model that reads word images in deep convolutional sequences. The new model is referred as Deep-Text Recurrent Network (DTRN), of which the pipeline is shown in Fig.~\ref{fig:pipeline}. It takes both the advantages of the deep CNN for image representation learning and the RNN model for sequence labelling, with the following appealing properties:

\noindent\textbf{1) Strong and high-level representation without character segmentation} --
The DTRN generates a convolutional image sequence, which is explicitly ordered by scanning a sliding window through a word image. The CNN sequence captures meaningful high-level representation that is robust to various image distortions. It differs significantly from manually-designed sequential features used by most prior studies based on sequence labelling~\cite{Breuel2013,Graves2009,Su2014}. The sequence is generated without any low-level operation or challenging character segmentation.

\noindent\textbf{2) Exploiting context information}
In contrast to existing systems~\cite{Bissacco2013,Jaderberg2014,Wang2012} that read each character independently, we formulate this task as a sequence labelling problem.
Specifically, we build our system on the LSTM, so as to capture the interdependencies inherent in the deep sequences. Such consideration allows our system to recognize highly ambiguous words, and work reliably without either pre- or post-processing. In addition, the recurrence allows it to process sequences of various lengths, going beyond traditional neural networks of fixed-length input and output.

\noindent\textbf{3) Process unknown words and arbitrary strings}
With properly learned deep CNNs and RNNs, our model does not depend on any pre-defined dictionary, unlike exiting studies~\cite{Jaderberg2015b,Jaderberg2014,Wang2012}, and  it can process unknown words, and arbitrary strings, including multiple words.

%We note that CNN and RNN have been widely used in many vision tasks.
We note that CNN and RNN have been independently exploited in the domain of text recognition. Our main contribution in this study is to develop a unified deep recurrent system that leverages both the advantages of CNN and RNN for the difficult scene text recognition problem, which has been solved based on analyzing character independently. This is the first attempt to show the effectiveness of exploiting convolutional sequence with sequence labeling model for this challenging task. We highlight the considerations required to make this system reliable and discuss the unique advantages offered by it.
The proposed DTRN demonstrate promising results on a number of benchmarks, improving recent results of \cite{Jaderberg2014,Almazan2014} considerably.

\section{Related Work}
%Word image recognition has gained increasing attention over the last several years.
Previous work mainly focuses on developing a powerful character classifier with manually-designed image features.
A HoG feature with random ferns was developed for character classification in \cite{Wang2011}. Neumann and Matas proposed new oriented strokes for character detection and classification \cite{Neumann2013}.
Their performance is limited by the low-level features. In \cite{Lee2014}, a mid-level representation of characters was developed by proposing a discriminative feature pooling. Similarly, Yao \emph{et al.} proposed the mid-level Strokelets to describe the parts of characters \cite{Yao2014}.

%then a whole character is represented by a bag of strokelets with an additional HoG feature

Recent advances of DNN for image representation encourage the development of more powerful character classifiers, leading to the state-of-the-art performance on this task. The pioneer work was done by LeCun \emph{et al.}, who designed a CNN for isolated handwriting digit recognition \cite{LeCun1998}. A two-layer CNN system was proposed for both character detection and classification in \cite{Wang2012}. PhotoOCR system employs a five-layer DNN for character recognition \cite{Bissacco2013}. Similarly, Jaderberg \emph{et al.} \cite{Jaderberg2014} proposed novel deep features by employing a Maxout CNN model for learning common features, which were subsequently used for a number of different tasks, such as character classification, location optimization and language model learning.

These approaches treat isolated character classification and subsequent word recognition separately. They do not unleash the full potential of word context information in the recognition. They often design complicated optimization algorithm to infer word string by incorporating multiple additional visual cues, or require a number of post-processing steps to refine the results \cite{Jaderberg2014,Bissacco2013}. Our model differs significantly from them by exploring the recurrence of deep features, allowing it to leverage the underlying context information to directly recognise the whole word image in a deep sequence, without a language model and any kind of post-processing.

% This makes it more straightforward that without any post-processing step.

There is another group of studies that recognise text strings from the whole word images. Almazan \emph{et al.} \cite{Almazan2014} proposed a subspace regression method to jointly embed both word image and its string into a common subspace.
%where word is recognised by simply computing the nearest neighbour.
 A powerful CNN model was developed to compute a deep feature from a whole word image in \cite{Jaderberg2015b}. Again, our model differs from these studies in the deep recurrent nature. Our  sequential feature includes explicit spatial order information, which is crucial to discriminate the order-sensitive word string. While the other global representation would lost such strict order, leading to poorer discrimination power. Furthermore, the model of \cite{Jaderberg2015b} is strictly constrained by the pre-defined dictionary, making it unable to recognise a novel word. By contrast, our model can process an unknown word. %, providing a more principled approach that generalizes better.

For unconstrained recognition, Jaderberg \emph{et al.} proposed another CNN model, which incorporates a Conditional Random Field \cite{Jaderberg2015a}. This model recognizes word strings in character sequences, allowing it for processing a single unknown word. But the model is highly sensitive to the non-character space, making it difficult to recognize multiple words.
%The sequential character classifiers stop until they output a non-character label or meet a maximum number of 23 characters. This limitation makes it impossible to recognize multiple words in a text-line, which are often separated by the non-character spaces, or the strings having more than 23 characters.
% Our recurrent model does not have such limitations, so is a more flexible approach that generalizes better.
 Our recurrent model can process arbitrary strings, including multiple words, and thus generalizes better.
 Our method also relates to \cite{Su2014}, where a RNN is built upon HOG features. However, its performance is significantly limited by the HOG.
%  which is highly sensitive to multiple low-level image distortions.
 % While the strong robustness of our deep CNN sequence is crucial to the success of our model.
  While the strong deep CNN feature is crucial to the success of our model.

Our approach is partially motivated by the recent success of deep models for image captioning, where the combination of the CNN and RNN has been applied~\cite{Andrej2015,Donahue2014,Alsharif2013}. They explored the CNN for computing a deep feature from a whole image, followed by a RNN to decode it into a sequence of words. ReNet \cite{Visin2015} was proposed to directly compute the deep image feature by using four RNN to sweep across the image. Generally, these models do not explicitly store the strict spatial information by using the global image representation. By contrast, our word images include explicit order information of its string, which is a crucial cue to discriminate a word.
%While this order information is not included explicitly in the global image representation of the image captation.
%For the image caption, the image and its target captions do not have such strictly spatial correlation. Global representation may include implicit spatial information of the objects in an image, but applying it to the word image would significantly loss the strict order information.
%Recently, ReNet \cite{Visin2015} was proposed to directly compute the deep image feature by using four RNN to sweep across the image. Similarly, it may not store the explicit order information by encoding the image into a single feature vector.
Our goal here is to derive a set of robust sequential features from the word image, and design an new model that bridges the image representation learning and sequence labelling task.

%Goodfellow \emph{et al.} developed a street number recognition system by using a CNN with multiple position-sensitive character classifier outputs \cite{}.

%our method%%%%%%%%%%%%%%%%%%%%%%%%%%%%%%%%%%%%%%%%%%%%%%%%%%%%%%%%%%%%%%%%%%%%%%%%%%%%%%%%%%%%%%%%%%%%%%%%%%%
\section{Deep-Text Recurrent Networks}
%This section presents details of the proposed deep-text recurrent networks (DTRN).

%Recent advances in machine translation, handwriting, and speech recognition show that the unrolling property of the RNN model makes it highly flexible to transfer sequence of varying lengths into different domains. This property is applied in machine translation for both encoding and decoding processes \cite{Sutskever2014}. An input transcript from a source language is encoded into a compact sharing feature by a RNN encoder, while another RNN model decodes this feature into a transcript of a target language.

The pipeline of Deep-Text Recurrent Network (DTRN) is shown in Fig.~\ref{fig:pipeline}. It starts by encoding a given word image into an ordered sequence with a specially designed CNN. Then a RNN is employed to decode (recognise) the CNN sequence into a word string. The system is end-to-end, i.e. it takes a word image as input and directly outputs the corresponding word string, without any pre- and post-processing steps. Both the input word image and output string can be of varying lengths.
This section revisits some important details of CNN and RNN and highlight the considerations that make their combination reliable for scene text recognition.

%The proposed DTRN works in a similar pipeline by encoding the word image into an . Then a RNN system is proposed to   which is input with a word image and outputs corresponding word string,

\begin{figure*}
%\abovecaptionskip
%\belowcaptionskip
\centering
\includegraphics[height=4cm,width=15cm]{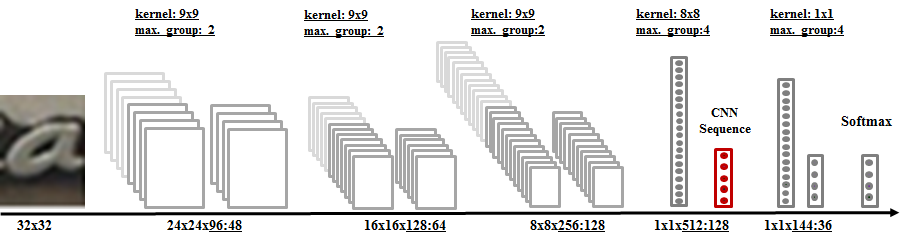}
%\qquad
%\includegraphics[height=4cm,width=5.5cm]{figures/rnn.png}
%\vskip -0.2cm
%\subfigure[Maxout CNN]{\includegraphics[height=3.5cm,width=11.5cm]{figures/CNN1.png}}\qquad
%\subfigure[RNN]{\includegraphics[height=3.5cm,width=5.5cm]{figures/DTRN.png}}

   \caption{The structures of our maxout CNN model.}
\label{fig:maxoutCNN}
\end{figure*}

Formally, we formulate the word image recognition as a sequence labeling problem. We maximize the probability of the correct word strings ($S_w$), given an input image ($I$),
 \begin{eqnarray}
\abovedisplayskip
\belowdisplayskip
\hat{\theta}=\arg\max_\theta\sum_{(I, S_w)}\log P(S_w|I;\theta),
\end{eqnarray}
where $\theta$ are the parameters of the recurrent system. $(I, S_w)\in\Omega$ is a sample pair from a training set, $\Omega$, where $S_w=\{S_w^1,S_w^2,...,S_w^K\}$ is the ground truth word string (containing $K$ characters) of the image $I$. Commonly, the chain rule is applied to model the joint probability over $S_w$,
 \begin{eqnarray}
\abovedisplayskip
\belowdisplayskip
\log P(S_w|I;\theta)=\sum_{i=1}^K\log P(S_w^i|I,S_w^0,...,S_w^{i-1};\theta)
\end{eqnarray}

Thus we optimize the sum of the log probabilities over all sample pairs in the training set ($\Omega$) to learn the model parameters. We develop a RNN to model the sequential probabilities $P(S_w^i|I,S_w^0,...,S_w^{i-1})$, where the variable number of the  sequentially conditioned characters can be expressed by an internal state of the RNN in hidden layer, $h_t$. This internal state is updated when the next sequential input $x_t$ is presented by computing a non-linear function $\mathcal{H}$,
\begin{eqnarray}
\abovedisplayskip
\belowdisplayskip
h_{t+1}=\mathcal{H}(h_t,x_t)
\end{eqnarray}
where the non-linear function $\mathcal{H}$ defines exact form of the proposed recurrent system. $X=\{x_1,x_2,x_3,...,x_T\}$ is the sequential CNN features computed from the word image,
\begin{eqnarray}
\abovedisplayskip
\belowdisplayskip
\{x_1,x_2,x_3,...,x_T\}=\varphi(I)
\end{eqnarray}

Designs of the $\varphi$ and $\mathcal{H}$ play crucial roles in the proposed system. We develop a CNN model to generate the sequential $x_t$, and define $\mathcal{H}$ with a long short-term memory (LSTM) architecture~\cite{Hochreiter1997}.

% which has achieved excellent performance on a number of sequence modelling tasks, such as language translation \cite {Sutskever2014}, handwriting \cite{Graves2009}, and speech recognition \cite{Graves2014}.

\subsection{Sequence Generation with Maxout CNN}

The main challenge of obtaining low-level sequential representation from the word images arises from the difficulties of correct segmentation at either pixel or character level.
%
%This makes it extremely difficult to manually design a reliable hand-craft feature. Deep CNN models have recently achieved great success in a number of vision tasks based on its high-level representation, such as for image classification \cite{Krizhevsky2012} and object detection \cite{Girshick2014}. Thus it is natural to develop a powerful CNN model to compute the high-level features from a word image, enabling it with strong robustness against the complicated background affects and multiple low-level image distortions. Also differing from most previous work that apply the CNN model to extract a global feature from the whole image,
%
We argue that it is not necessary to perform such low-level feature extraction.
On the contrary, it is more natural to describe word strings in sequences where their explicit order information is retained. This information is extremely important to discriminate a word string. Furthermore, the variations between continuous examples in a sequence should encode additional information, which could be useful in making more reliable prediction. By considering these factors, we propose to generate an explicitly ordered deep sequence with a CNN model, by sliding a sub window through the word image.

To this end, we develop a Maxout network~\cite{Goodfellow2013} for computing the deep feature. It has been shown that the Maxout CNN is powerful for character classification \cite{Jaderberg2014,Alsharif2013}. The basic pipeline is to compute point-wise maximum through a number of grouped feature maps or channels. Our networks is shown in Fig~\ref{fig:maxoutCNN} (a), the input image is of size $32 \times 32$, corresponding to the size of sliding-window. It has five convolutional layers, each of which is followed by a two- or four-group Maxout operation, with various numbers of feature maps, i.e.\ 48, 64, 128,128 and 36, respectively. Similar to the CNN used in \cite{Jaderberg2014}, our networks does not involve any pooling operation, and the output of last two convolutional layers are just one pixel. This allows our CNN to convolute the whole word images at once, leading to a significant computational efficiency. For each word image, we resize it into the  height of 32, and keep its original aspect ratio unchanged. We apply the learned filters to the resized image, and get a 128D CNN sequence directly from the output of last second convolutional layer. This operation is similar to computing deep feature independently from the sliding-window by moving it densely through the image, but with much computational efficiency. Our Maxout CNN is trained on 36-class case insensitive character images.

% This allows the CNN to learn multiple modes of the data by computing the maximum response over a mixture of linear models. The main structure of our maxout CNN is presented in Fig.~\ref{fig:maxoutCNN}

%where the generated feature $x_t\in\mathbb{R}^{128}$, is of 128 dimensions, corresponding to the number of neurons in the last convolutional layer of our maxout CNN, as shown in Figure x. $T$ is the sequence length varied in different word images.

\subsection{Sequence Labeling with RNN}
\begin{figure}
\begin{center}
\includegraphics[height=5cm,width=8.5cm]{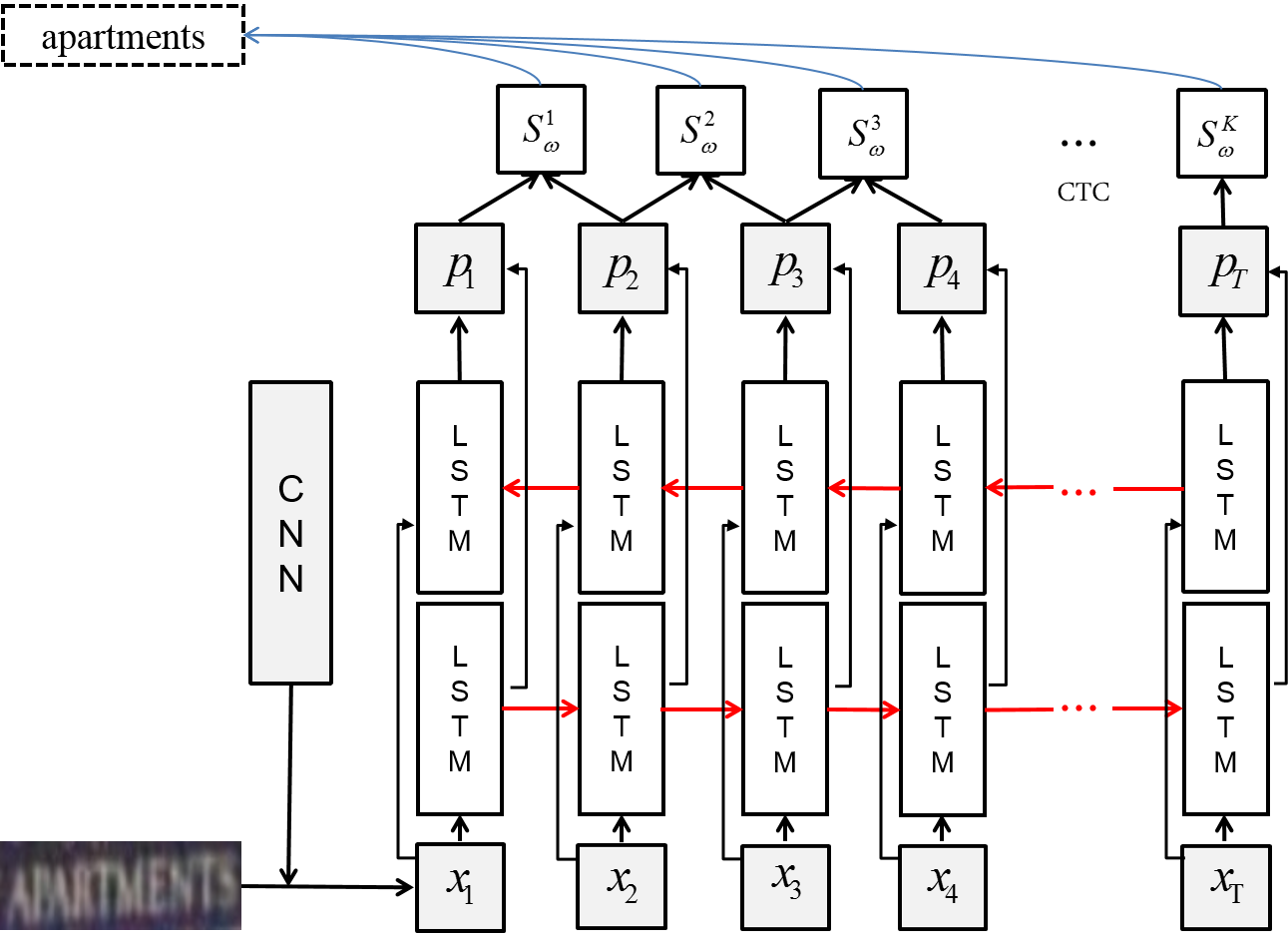}
\end{center}
\vskip -0.2cm
   \caption{The structure of our recurrent neural networks.}
\label{fig:rnn}
\end{figure}

We believe that the interdependencies between the convolutional sequence include meaningful context information which would be greatly helpful to identify an ambitious character.
RNN has shown strong capability for learning meaningful structure from an ordered sequence.
Another important property of the RNN is that the rate of changes of the internal state can be finely modulated by the recurrent weights, which contributes to its robustness against localised distortions of the input data~\cite{Graves2009}.
Thus we propose the use of RNN in our framework to model the generated CNN sequence $\{x_1, x_2, x_3,...,x_T\}$. The structure of our RNN model is shown in Fig.~\ref{fig:rnn}.

The main shortcoming of the standard RNN is the vanishing gradient problem, making it hard to transmit the gradient information consistently over long time \cite{Hochreiter1997}. This is a crucial issue in designing a RNN model, and the long short-term memory (LSTM) was proposed specially to address this problem \cite{Hochreiter1997}. The LSTM defines a new neuron or cell structure in the hidden layer with three additional multiplicative gates: the \emph{input gate}, \emph{forget gate} and \emph{output gate}. These new cells are referred as memory cells, which allow the LSTM to learn meaningful long-range interdependencies.
We skip standard descriptions of the LSTM memory cells and its formulation, by leaving them in the supplementary material.
%The structure of the memory cells is described in Fig.~\ref{fig:lstm}. We skip the standard formulation of LSTM and leave it in the supplementary material.

%The sequence labelling of varying lengths is processed by recurrently implementing the LSTM memory for each sequential input $x_t$, such that all LSTMs share the same parameters. The output of the LSTM $h_t$ is used to fed to the LSTM at next input $x_{t+1}$. It is also used to compute the current output, which is transformed to the estimated probabilities over all possible characters. It finally generates a sequence of the estimations with the same length of input sequence, $\textbf{p}=\{p_1,p_2,p_3,...,p_T\}$.

\begin{figure*}
%\abovecaptionskip
%\belowcaptionskip
\centering
%\begin{center}
%\includegraphics[height=3.5cm,width=11cm]{figures/training_process.png}
\subfigure[the LSTM output]{\includegraphics[height=3.5cm,width=3cm]{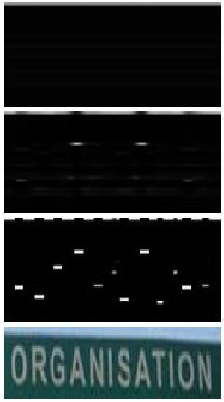}}
\subfigure[the CTC path]{\includegraphics[height=3.5cm,width=3cm]{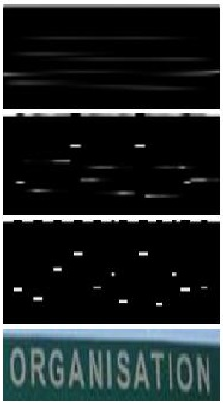}}
\subfigure[$\textbf{p}$ and $\pi$ ]{\includegraphics[height=3.5cm,width=5cm]{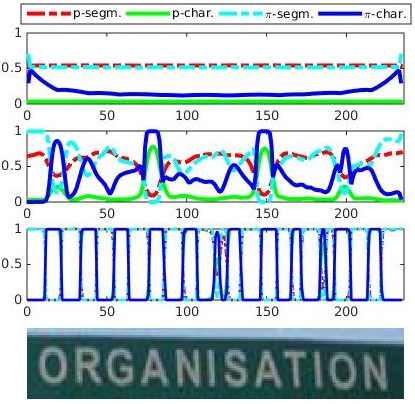}}
\subfigure[DTRN vs DeepFeatures]{\includegraphics[height=3.5cm,width=6cm]{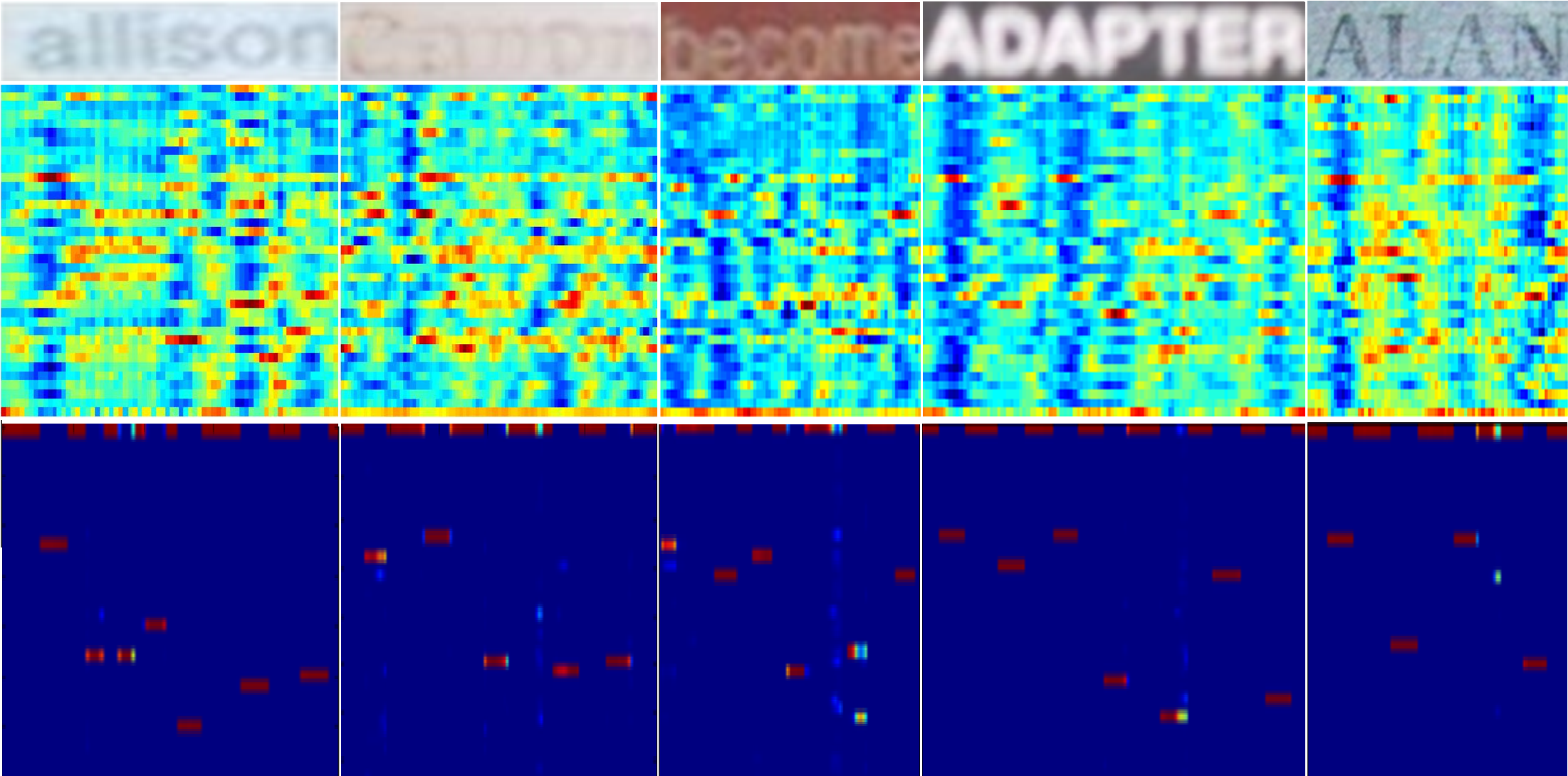}}
%\end{center}
\caption{(a-c)RNNs training process recorded at epoch 0 (row 1), 5 (row 2) and 50 (row 3) with a same word image (row 4). (a) the LSTM output ($\textbf{p}$); (b) the CTC path ($\pi$) mapped from ground truth word string ($\mathcal{B}^{-1}(S_w)$); (c) maximum probabilities of the character and segmentation line with $\textbf{p}$ and $\pi$; (d) output confident maps of the DeepFeatures (middle) and the LSTM layer of the DTRN (bottom).}
\label{fig:train_lstm}
\end{figure*}

The sequence labelling of varying lengths is processed by recurrently implementing the LSTM memory for each sequential input $x_t$, such that all LSTMs share the same parameters. The output of the LSTM $h_t$ is fed to the LSTM at next input $x_{t+1}$. It is also used to compute the current output, which is transformed to the estimated probabilities over all possible characters. It finally generates a sequence of the estimations with the same length of input sequence, $\textbf{p}=\{p_1,p_2,p_3,...,p_T\}$.

%(36 case-insensitive with an additional non-character labels in our case )

Due to the unsegmented nature of the word image at the character level, the length of the LSTM outputs ($T$) is not consistent with the length of a target word string, $|S_w|=K$. This makes it difficult to train our recurrent system directly with the target strings. To this end, we follow the recurrent system developed for the handwriting recognition \cite{Graves2009} by applying a connectionist temporal classification (CTC) \cite{Graves2005} to approximately map the LSTM sequential output ($\textbf{p}$) into its target string as follow,
\begin{eqnarray}
%\abovedisplayskip
%\belowdisplayskip
S_w^\star\approx\mathcal{B}(\arg\max_\pi P(\pi|\textbf{p}))
\end{eqnarray}
where the projection $\mathcal{B}$ removes the repeated labels and the non-character labels \cite{Graves2005}. For example, $\mathcal{B}(-gg-o-oo-dd-)=good$. The CTC looks for an approximately optimized path ($\pi$) with maximum probability through the LSTMs output sequence, which aligns the different lengths of LSTM sequence and the word string.

%\begin{figure}
%\begin{center}
%\includegraphics[width=0.75\linewidth]{figures/lstm.jpg}
%\end{center}
%\vskip -0.2cm
%   \caption{LSTM memory cell.}
%\label{fig:lstm}
%\end{figure}

The CTC is specifically designed for the sequence labelling tasks where it is hard to  pre-segment the input sequence to the segments that exactly match a target sequence. In our RNN model, the CTC layer is directly connected to the outputs of LSTMs, and works as the output layer of the whole RNN. It not only allows our model to avoid a number of complicated post-processing (e.g.\ transforming the LSTM output sequence into a word string), but also makes it possible to be trained in an end-to-end fashion by minimizing an overall loss function over $(X,S_w)\in\Omega$. The loss for each sample pair is computed as sum of the negative log likelihood of the true word string,
\begin{eqnarray}
%\abovedisplayskip
%\belowdisplayskip
L(X,S_w)=-\sum_{i=1}^K \log P(S_w^i|X)
\end{eqnarray}

Finally, our RNNs model follows a bidirectional LSTM architecture, as shown in Fig.~\ref{fig:rnn} (b). It has two separate LSTM hidden layers that process the input sequence forward and backward, respectively. Both hidden layers are connected to the same output layers, allowing it to access both past and future information. In several sequence labelling tasks, such as handwriting recognition \cite{Graves2009} and phoneme recognition \cite{Graves2005}, the bidirectional RNNs have shown stronger capability than the standard RNNs. Our RNNs model is trained with the Forward-Backward Algorithm that jointly optimizes the bidirectional LSTM and CTC. Details are presented in the supplementary material.

\subsection{Implementation Details}
Our CNN model is trained on about $1.8 \times 10^5$ character images cropped from the training sets of a number of benchmarks by \cite{Jaderberg2014}. We generate the CNN sequence by applying the trained CNN with a sliding-window, followed by a column-wise normalization.
Our recurrent model contains a bidirectional LSTM. Each LSTM layer has 128  cell memory blocks. The input layer has 128 neurons (corresponding to 128D CNN sequence),  which are fully connected to both hidden layers. The outputs of two hidden layers are concatenated, and then fully connected to the output layer of LSTM with 37 output classes (including the non-character), by using a softmax function.
%, as shown in Fig.~\ref{fig:maxout_vs_lstm} (bottom).
%
Our RNN model has 273K parameters in total. In our experiments, we found that adding more layers LSTM does not lead to better results in our task. We conjecture that LSTM needs not be deep, given the deep CNN which has provided strong representations.
%which are initialized with a Gaussian distribution of mean 0 and standard deviation 0.01 in the training process.

The recurrent model is trained with steepest descent. The parameters are updated per training sequence by using a learning rate of $10^{-4}$ and a momentum of 0.9.
%Each input sequence is randomly selected from the training set,
We perform forward-backward algorithm \cite{Graves2006} to jointly optimize the LSTM and CTC parameters, where a forward propagation is implemented through whole network, followed by a forward-backward algorithm that aligns the ground truth word strings to the LSTM outputs, $\pi \in \mathcal{B}^{-1}(S_w)$, $\pi,\textbf{p}\in\mathbb{R}^{37\times T}$. The loss of E.q.(6) is computed approximately as:
\begin{eqnarray}
L(X,S_w)\approx -\sum_{t=1}^T\log P(\pi_t|X)
\end{eqnarray}
Finally, the approximated error is  propagated backward to update the parameters. The RNN is trained on about $3000$ word images (all characters of them are included in previously-used $1.8 \times 10^5$ character images), taken from the training sets of three benchmarks used bellow. The training process is shown in Fig.~\ref{fig:train_lstm}.

\section{Experiments and Results}

%The performance of the proposed DTRN is compared against start-of-the-art methods
The experiments were conducted
on three standard benchmarks for cropped word image recognition: the Street View Text ($SVT$) \cite{Wang2011}, ICDAR 2003 ($IC03$) \cite{Lucas2003} and IIIT 5K-word ($IIIT5K$) \cite{Mishra2012}.
%
%
%\textbf{Preprocessing}.Each cropped image is converted into gray scale and normalized by subtracting the image mean and dividing by the standard deviation.For the later CNN sequence,each column is normalized by diving by the square root of one which is the sum of squares of column elements.This normalization is reliable considering the sigmoid and tangent function existed in each LSTM cell.
%
The $SVT$ has 647 word images collected from Google Street View of road-side scenes.
%The images are highly challenging with large variations in illumination, character sizes, and fonts. Some images are significantly blurred.
It provides a lexicon of 50 words per image for recognition (SVT-50). The $IC03$ contains 860 word images cropped from 251 natural images.
Lexicons with 50 words per image (IC03-50) and all words of the test set (IC03-FULL)  are provided. The $IIIT5K$ is comprised of 5000 cropped word images from both scene and born-digital images. The dataset is split into subsets of 2000 and 3000 images for training and test. Each image is associated with lexicons of 50  (IIIT5k-50) and 1k words (IIIT5k-1k) for test.

\subsection{DTRN vs DeepFeatures}
\label{sec:dtrn_vs_maxout}

\begin{figure*}
\begin{center}
\includegraphics[height=7cm,width=16cm]{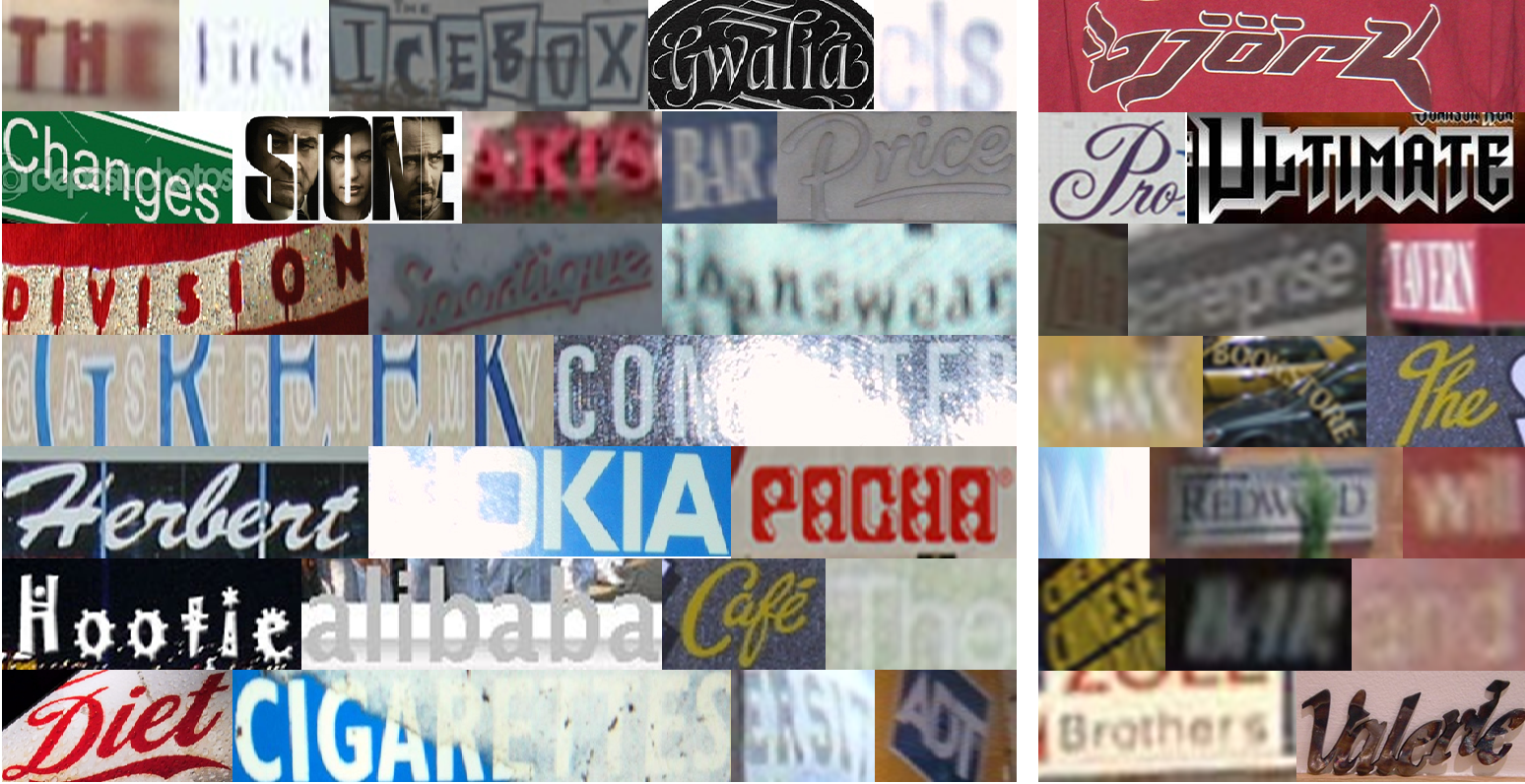}
\end{center}
\vskip -0.1cm
   \caption{(Left) Correct recognitions; (Right) Incorrect samples.}
\label{fig:result_examples}
\end{figure*}

%\begin{figure}[t]
%\begin{center}
%\includegraphics[width=8.5cm]{figures/maxout_vs_lstm.png}
%\end{center}
%   \caption{Output confidence maps of the Maxout CNN (middle) and the LSTM output layer of the DTRN (bottom) on a number of ambiguous images (top).}
%\label{fig:maxout_vs_lstm}
%\end{figure}

The recurrence property of the DTRN makes it distinct against the current deep CNN models, such as DeepFeatures \cite{Jaderberg2014}) and the system of \cite{Wang2012}. The advantage is shown clearly in Fig.~\ref{fig:train_lstm} (d), where the output maps of the LSTM layer and the Maxout CNN of DeepFeatures are compared. As can be observed, our maps are much clearer than those of the DeepFeatures in a number of highly ambiguous word images.
The character probability distribution and segmentation are shown accurately on our maps, indicating the excellent capability of our model for correctly identifying word texts from challenging images. The final word recognition is straightforward by simply applying the $\mathcal{B}$ projection (E.q. 5) on these maps.
However, the maps of DeepFeatures are highly confused, making it extremely difficult to infer the correct word strings from their maps.
%It requires an optimization function to incorporate multiple additional cues, followed by a number of complicated post-processing steps to further refine the results.
Essentially, the recurrent property of DTRN allows it to identify a character robustly from a number of continuous regions or sided windows, while the DeepFeatures classifies each isolated region independently so that it is confused when a located region just includes a part of the character or multiple characters.

%Thus it require a complicated optimization function to incorporate multiple additional cues, which are hard to
%
%A complicated optimization function was designed to incorporate multiple additional cues, makes the system difficult to jointly optimize with all these additional information, and also requires to tune a number of model parameters manually. Furthermore, some of these additional cues are even hard to learn correctly and reliably.
%
%
%The DeepFeatures compute word strings directly from these highly ambiguous maps by designing a complicated optimization function to incorporate multiple additional cues, such as both case sensitive and insensitive maps, a learned language model, a pre-defined lexicon, breakpoint locations, and a number of heuristic cues. This makes the system difficult to jointly optimize with all these additional information, and also requires to tune a number of model parameters manually. Furthermore, some of these additional cues are even hard to learn correctly and reliably.

\subsection{Comparisons with State-of-the-Art}
 \begin{table*}[tb]
%\captionsetup{font={footnotesize}}
%\centering
%\footnotesize

%\begin{center}
\begin{multicols}{2}
\begin{minipage}[t]{0\textwidth}
\footnotesize
%\begin{minipage}[l]{1.2\linewidth}

%\centering
%
%\begin{tabular}{l|c|c|c|c|c}
\begin{tabular}{|*{6}{l|c|c|c|c|c|}}

%\hline
%\multirow{3}{*}{Method}
%&\multicolumn{5}{|c}{Cropped Word Recognition Accuracy(\%)} \\
%\cline{2-6}
%&IC03 &IC03 &SVT&IIIT5k&IIIT5k\\
%&-50  &-FULL&-50&-50&-1K\\

\hline
\multirow{2}{*}{Method}
&\multicolumn{5}{|c|}{Cropped Word Recognition Accuracy(\%)} \\
\cline{2-6}
%&\multicolumn{1}{|c|}{IC03-50}
%&\multicolumn{1}{|c|}{IC03-FULL}
%&\multicolumn{1}{|c|}{SVT-50}
%&\multicolumn{1}{|c|}{IIIT5k-50}
%&\multicolumn{1}{|c|}{IIIT5k-1K}\\
&IC03-50
&IC03-FULL
&SVT-50
&IIIT5k-50
&IIIT5k-1K\\

%&\multicolumn{1}{|c|}{IC03-50}
%&\multicolumn{1}{|c|}{IC03-FULL}
%&\multicolumn{1}{|c|}{SVT-50}
%&\multicolumn{1}{|c|}{IIIT5k-50}
%&\multicolumn{1}{|c|}{IIIT5k-1K}\\
\hline
\hline
%ABBYY\cite{Wang2011} & 56.0 & 55.0 & 35.0 & 24.3 & -\\
%Wang et al. 2011 & 76.0 & 62.0 & 57.0 & 64.1 & 57.5\\
%Mishra et al. 2012 & 81.8 & 67.8 & 73.2 & - & -\\
%%Novikova et al. 2012 & 82.8 & - & 72.9 & - & -\\
%%Shi et al. 2013 & 87.4 & 79.3 & 73.5 & - & -\\
%\citeauthor{Shi2013} 2013 & 87.4 & 79.3 & 73.5 & - & -\\
%Lee et al. 2014 & 88.0 & 76.0 & 80.0 & - & -\\
%Yao et al. 2014 & 88.5 & 80.3 & 75.9 & 80.2 & 69.3\\
%\hline
%Wang et al. 2012 & 90.0 & 84.0 & 70.0 & - & -\\
%Alsharif et al. 2013 & 93.1 & 88.6 & 74.3 & - & -\\
%Su and Lu 2014 &92.0&82.0&83.0&-&-\\
%DeepFeatures  & 96.2 & 91.5 & 86.1 & - & -\\
%\hline
%\citeauthor{Goel2013} 2013 & 89.7 & - & 77.3 & - & -\\
%Almaz$\acute{a}$n et al. 2014 & - & - & 87.0 & 88.6 & 75.6\\
%\hline
Wang et al. 2011 & 76.0 & 62.0 & 57.0 & 64.1 & 57.5\\
Mishra et al. 2012 & 81.8 & 67.8 & 73.2 & - & -\\
Novikova et al. 2012 & 82.8 & - & 72.9 & - & -\\
TSM+CRF\cite{Shi2013} & 87.4 & 79.3 & 73.5 & - & -\\
Lee et al. 2014 & 88.0 & 76.0 & 80.0 & - & -\\
Strokelets\cite{Yao2014} & 88.5 & 80.3 & 75.9 & 80.2 & 69.3\\
\hline
Wang et al. 2012 & 90.0 & 84.0 & 70.0 & - & -\\
Alsharif and Pineau 2013 & 93.1 & 88.6 & 74.3 & - & -\\
Su and Lu 2014 &92.0&82.0&83.0&-&-\\
DeepFeatures & 96.2 & 91.5 & 86.1 & - & -\\
\hline
Goel et al. 2013 & 89.7 & - & 77.3 & - & -\\
Almaz$\acute{a}$n et al. 2014 & - & - & 87.0 & 88.6 & 75.6\\
\hline

DTRN & \textbf{97.0} & \textbf{93.8} & \textbf{93.5} & \textbf{94.0} & \textbf{91.5} \\
\hline
\hline
%&\multicolumn{5}{|c|}{Training on Large Additional Datasets} \\
%\cline{2-6}
PhotoOCR & - & - & 90.4 & - & -\\
%Jaderberg2014 & 96.7 & 94.0 & 92.6 & - & -\\

Jaderberg2015a & 97.8 & 97.0& 93.2 & 95.5 & 89.6\\
Jaderberg2015b & 98.7 & 98.6 & 95.4 & 97.1 & 92.7\\
\hline
\end{tabular}
\end{minipage}
%\vskip +0.3cm

%\begin{minipage}[l]{1.2\linewidth}
%\centering
%\includegraphics[height=2cm,width=4.5in]{figures/unconstrained.png}
%%\label{fig:figureX}
%\end{minipage}

\begin{minipage}[t]{0.65\textwidth}
\centering
\includegraphics[width=2in,height=2.6in]{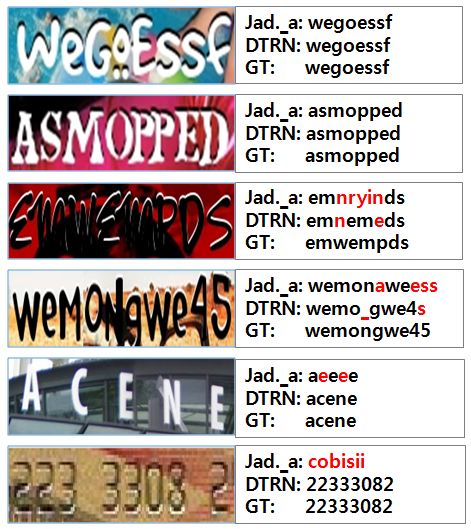}
%\label{fig:figureX}
\end{minipage}

\end{multicols}
 \caption{Cropped word recognition results on the SVT, ICDAR 2003, and IIIT 5K-word. %$Jaderberg2014$ corresponds to $CHAR+2$ model in \cite{Jaderberg2014b}, and $Jaderberg2015a$ is extended from the $Jaderberg2014$ by incorporating a CRF and a N-gram predictor.
 The bottom figure shows unconstrained recognitions of the DTRN and the publicly available model \cite{Jaderberg2014b}, which is similar to $Jaderberg2015a$. Obviously, it seems to be sensitive to non-character spaces.}\label{table:fullresult}
\end{table*}

The evaluation is conducted by following the standard protocol, %\cite{Jaderberg2014,Bissacco2013,Almazan2014,Wang2011,Wang2012},
where each word image is associated with a lexicon, and edit distance is computed to find the optimized word. The recognition results by the DTRN are presented in Fig.~\ref{fig:result_examples}, including both the correct and incorrect recognitions. As can been seen, the DTRN demonstrates excellent capability on recognising extremely ambiguous word images, some of which are even hard to human. This is mainly beneficial from its strong ability to leverage explicit order and meaningful word context information. The results on three benchmarks are compared with the state-of-the-art in Table~\ref{table:fullresult}.
%They are compared systematically to recent results achieved by mid-level features, deep neural networks, and  whole image based representation, along with discussions on their advantages and disadvantages.

%The results of the DTRN on the SVT, IC03 and  IIIT5K are presented in Table~\ref{table:fullresult}. They are compared systematically to recent results achieved by mid-level features \cite{Yao2014, Lee2014,Shi2013,Novikova2012,Mishra2012}, deep neural networks \cite{Jaderberg2014,Alsharif2013,Wang2012,Su2014} and  whole image based representation \cite{Almazan2014,Goel2013,Jaderberg2015a,Jaderberg2015b}, along with discussions on their advantages and disadvantages.

\textbf{Mid-level representation}: Strokelet \cite{Yao2014} and Lee \emph{et al.}'s method \cite{Lee2014} achieved leading performance based on the mid-level features. Though they show large improvements over conventional low-level features, their performance are not comparable to ours, with significant reductions in accuracies in all the three datasets.

\textbf{Deep neural networks}: As shown in Table 1, the DNN methods largely outperform the mid-level approaches, with close to $10\%$ of improvement in all cases. The considerable performance gains mainly come from its ability to learn a deep high-level feature from the word image. Su and Lu's method obtained accuracy of $83\%$ on SVT by building a RNN model upon the HOG features. DeepFeatures achieved leading results on both the SVT and IC03 datasets. However, the DeepFeatures are still built on isolate character classifier. By training a similar CNN model with the same training data, the DTRN achieved significant improvements over the DeepFeatures in all datasets. The results agree with our analysis conducted above. On the widely-used SVT, our model outperforms the DeepFeatures considerably from $86.1\%$ to $93.5\%$, indicating the superiority of our recurrent model in connecting the isolated deep features sequentially for recognition.  Furthermore, our system does not need to learn the additional language model and character location information, all of which are optimized jointly and automatically by our RNN in an end-to-end fashion.

\textbf{Whole image representation}:
%The DTRN is compared with Almazan \emph{et al.}'s approach \cite{Almazan2014} based on the whole word image representation.
Almazan \emph{et al.}'s approach, based on the whole word image representation, achieved $87.0\%$ accuracy on the SVT \cite{Almazan2014}, slightly over that of DeepFeatures. In the IIIT5k, it yielded $88.6\%$ and $75.6\%$ on small and large lexicons, surpassing previous results with a large margin. Our DTRN strives for a further step by reaching the accuracies of $94\%$ and $91.5\%$ on the IIIT5k. The large improvements may benefit from the explicit order information included in our CNN sequence. It is the key to increase discriminative power of our model for word representation, which is highly sensitive to the order of characters. The strong discriminative power can be further verified by the consistent high-performance of our system along with the increase of lexicon sizes, where the accuracy of Almazan \emph{et al.}'s approach drops significantly.

\textbf{Training on additional large datasets}: The PhotoOCR \cite{Bissacco2013} sets a strong baseline on the SVT ($90.4\%$) by using large additional training data.
 It employed about $10^7$ character examples to learn a powerful DNN classifier, and also trained a strong language model with a corpus of more than a trillion tokens. However, it involves a number of low-level techniques to over-segment characters, and jointly optimizes the segmentation, character classification and language model with beam search.  Furthermore, it also includes a number of post-processing steps to further improve the performance, making the system highly complicated.
 %such as punctuation search, secondary language scoring, shape model and junk filter, making the system highly complicated.
 The DTRN achieved $3.1\%$ improvement over the PhotoOCR, which is also significant by considering only a fraction of the training data (two orders of magnitude less data) we used. While our model works without a language model, and does not need any post-processing step.

Jaderberg \emph{et al.}  proposed several powerful deep CNN models by computing a deep feature from the whole word image \cite{Jaderberg2014b,Jaderberg2015a,Jaderberg2015b}. However, directly comparing our DTRN to these models may be difficult. First, these models was trained on $7.2\times10^6$ word images, comparing to ours $3\times10^3$ word images (with $1.8\times10^5$ characters). Nevertheless,
%it is worth pointing out that despite our model is trained with relatively smaller dataset,
our model achieves comparable results against $Jaderberg2015a$ with higher accuracies on the SVT and IIIT5k-1K. Importantly, the DTRN also provides unique capability for unconstrained recognition of any number of characters and/or word strings in a text-line. Several examples are presented in the figure of Table 1. $Jaderberg2015b$ model achieves the best results in all databases.  It casts the word recognition problem as a large-scale classification task by considering the images of a same word as a class.
%Thus the output classes in output layer of the CNN model should include all possible words, which should be at least equal to the total number of words in a regular English dictionary (about 90K). This causes a huge number of model parameters, making the model difficult to be trained.
Thus the output layer should include a large number of classes, e.g. 90,000, imposing a huge number of model parameters which are difficult to be trained.
%Furthermore, The $Jaderberg2015b$ is not flexible in that it cannot recognize a new word not trained.
Furthermore, it is not flexible to recognize a new word not trained.
While the scene texts often include many irregular word strings (the number could be unlimited) which are impossible to be known in advanced, such as "AB00d".
%Therefore their model needs to know all ground truth words in the test set a priori, and uses these word strings to generate corresponding synthetic word images for training.
Thus our DTRN can process unknown words and arbitrary strings, providing a more flexible approach for this task.

%\begin{minipage}[c]{0.2\linewidth}
%\includegraphics[width=2.0in]{figures/unconstrained.png}
%%\label{fig:figureX}
%\end{minipage}
\section{Conclusion}

We have presented a Deep-Text Recurrent Network (DTRN) for scene text recognition.  It models the task as a deep sequence labelling problem that overcomes a number of main limitations. It computes a set of explicitly-ordered deep features from the word image, which is not only robust to low-level image distortions, but also highly discriminative to word strings. The recurrence property makes it capable of recognising highly ambiguous images by leveraging meaningful word context information, and also allows it to process unknown words and arbitrary strings, providing a more principled approach for this task. Experimental results show that our model has achieved the state-of-the-art performance.

\section{ Acknowledgments}
This work is partly supported by National Natural Science Foundation of China (61503367, 91320101, 61472410), Guangdong Natural Science Foundation (2015A030310289), Guangdong Innovative Research Program (201001D0104648280, 2014B050505017) and Shenzhen Basic Research Program (KQCX2015033117354153). Yu Qiao is the corresponding author.

{
\bibliographystyle{aaai}
\bibliography{ref}

\begin{thebibliography}{}

\bibitem[\protect\citeauthoryear{Almaz{\'{a}}n \bgroup et al\mbox.\egroup
  }{2014}]{Almazan2014}
Almaz{\'{a}}n, J.; Gordo, A.; Forn{\'{e}}s, A.; and Valveny, E.
\newblock 2014.
\newblock Word spotting and recognition with embedded attributes.
\newblock {\em IEEE Trans. Pattern Analysis and Machine Intelligence (TPAMI)}
  36:2552--2566.

\bibitem[\protect\citeauthoryear{Alsharif and Pineau}{2013}]{Alsharif2013}
Alsharif, O., and Pineau, J.
\newblock 2013.
\newblock End-to-end text recognition with hybrid {HMM} maxout models.
\newblock {\em arXiv:1310.1811v1}.

\bibitem[\protect\citeauthoryear{Andrej and Li}{2015}]{Andrej2015}
Andrej, K., and Li, F.
\newblock 2015.
\newblock Deep visual-semantic alignments for generating image descriptions.
\newblock IEEE Computer Vision and Pattern Recognition (CVPR).

\bibitem[\protect\citeauthoryear{Bissacco \bgroup et al\mbox.\egroup
  }{2013}]{Bissacco2013}
Bissacco, A.; Cummins, M.; Netzer, Y.; and Neven, H.
\newblock 2013.
\newblock Photoocr: Reading text in uncontrolled conditions.
\newblock IEEE International Conference on Computer Vision (ICCV).

\bibitem[\protect\citeauthoryear{Breuel \bgroup et al\mbox.\egroup
  }{2013}]{Breuel2013}
Breuel, T.; UI-Hasan, A.; Azawi, M.; and Shafait, F.
\newblock 2013.
\newblock High-performance ocr for printed english and fraktur using lstm
  networks.
\newblock International Conference on Document Analysis and Recognition
  (ICDAR).

\bibitem[\protect\citeauthoryear{Donahue \bgroup et al\mbox.\egroup
  }{2015}]{Donahue2014}
Donahue, J.; Hendricks, L.; Guadarrama, S.; and Rohrbach, M.
\newblock 2015.
\newblock Long-term recurrent covolutional networks for visual recognition and
  description.
\newblock IEEE Computer Vision and Pattern Recognition (CVPR).

\bibitem[\protect\citeauthoryear{Goodfellow \bgroup et al\mbox.\egroup
  }{2013}]{Goodfellow2013}
Goodfellow, I.; Warde-Farley, D.; Mirza, M.; Courville, A.; and Bengio, Y.
\newblock 2013.
\newblock Maxout networks.
\newblock {\em arXiv:1302.4389v4}.

\bibitem[\protect\citeauthoryear{Graves and Jaitly}{2014}]{Graves2014}
Graves, A., and Jaitly, N.
\newblock 2014.
\newblock Towards end-to-end speech recognition with recurrent neural networks.
\newblock IEEE International Conference on Machine Learning (ICML).

\bibitem[\protect\citeauthoryear{Graves and Schmidhuber}{2005}]{Graves2005}
Graves, A., and Schmidhuber, J.
\newblock 2005.
\newblock Framewise phoneme classification with bidirectional lstm and other
  neural network architectures.
\newblock {\em Neural Networks} 18:602--610.

\bibitem[\protect\citeauthoryear{Graves and Schmidhuber}{2008}]{Graves2008}
Graves, A., and Schmidhuber, J.
\newblock 2008.
\newblock Offline handwriting recognition with multidimensional recurrent
  neural networks.
\newblock Neural Information Processing Systems (NIPS).

\bibitem[\protect\citeauthoryear{Graves \bgroup et al\mbox.\egroup
  }{2006}]{Graves2006}
Graves, A.; Fernandez, S.; Gomez, F.; and Schmidhuber, J.
\newblock 2006.
\newblock Connectionist temporal classification: Labelling unsegmented sequence
  data with recurrent neural networks.
\newblock IEEE International Conference on Machine Learning (ICML).

\bibitem[\protect\citeauthoryear{Graves, Liwicki, and
  Fernandez}{2009}]{Graves2009}
Graves, A.; Liwicki, M.; and Fernandez, S.
\newblock 2009.
\newblock A novel connectionist system for unconstrained handwriting
  recognition.
\newblock {\em IEEE Trans. Pattern Analysis and Machine Intelligence (TPAMI)}
  31:855--868.

\bibitem[\protect\citeauthoryear{He \bgroup et al\mbox.\egroup }{2015}]{He2015}
He, T.; Huang, W.; Qiao, Y.; and Yao, J.
\newblock 2015.
\newblock Text-attentional convolutional neural networks for scene text
  detection.
\newblock {\em arXiv:1510.03283}.

\bibitem[\protect\citeauthoryear{Hochreiter and
  Schmidhuber}{1997}]{Hochreiter1997}
Hochreiter, S., and Schmidhuber, J.
\newblock 1997.
\newblock Long short-term memory.
\newblock {\em Neural Computation} 9:1735--1780.

\bibitem[\protect\citeauthoryear{Huang \bgroup et al\mbox.\egroup
  }{2013}]{Huang2013}
Huang, W.; Lin, Z.; Yang, J.; and Wang, J.
\newblock 2013.
\newblock Text localization in natural images using stroke feature transform
  and text covariance descriptors.
\newblock IEEE International Conference on Computer Vision (ICCV).

\bibitem[\protect\citeauthoryear{Huang, Qiao, and Tang}{2014}]{Huang2014}
Huang, W.; Qiao, Y.; and Tang, X.
\newblock 2014.
\newblock Robust scene text detection with convolution neural network induced
  mser trees.
\newblock European Conference on Computer Vision (ECCV).

\bibitem[\protect\citeauthoryear{Jaderberg \bgroup et al\mbox.\egroup
  }{2014}]{Jaderberg2014b}
Jaderberg, M.; Simonyan, K.; Vedaldi, A.; and Zisserman, A.
\newblock 2014.
\newblock Synthetic data and artificial neural networks for natural scene text
  recognition.
\newblock Workshop in Neural Information Processing Systems (NIPS).

\bibitem[\protect\citeauthoryear{Jaderberg \bgroup et al\mbox.\egroup
  }{2015a}]{Jaderberg2015a}
Jaderberg, M.; Simonyan, K.; Vedaldi, A.; and Zisserman, A.
\newblock 2015a.
\newblock Deep structured output learning for unconstrained text recognition.
\newblock International Conference on Learning Representation (ICLR).

\bibitem[\protect\citeauthoryear{Jaderberg \bgroup et al\mbox.\egroup
  }{2015b}]{Jaderberg2015b}
Jaderberg, M.; Simonyan, K.; Vedaldi, A.; and Zisserman, A.
\newblock 2015b.
\newblock Reading text in the wild with convolutional neural networks.
\newblock {\em International Jounal of Computver Vision (IJCV)}.

\bibitem[\protect\citeauthoryear{Jaderberg, Vedaldi, and
  Zisserman}{2014}]{Jaderberg2014}
Jaderberg, M.; Vedaldi, A.; and Zisserman, A.
\newblock 2014.
\newblock Deep features for text spotting.
\newblock European Conference on Computer Vision (ECCV).

\bibitem[\protect\citeauthoryear{LeCun \bgroup et al\mbox.\egroup
  }{1998}]{LeCun1998}
LeCun, Y.; Bottou, L.; Bengio, Y.; and Haffner, P.
\newblock 1998.
\newblock Gradient-based learning applied to document recognition.
\newblock {\em Proceedings of the IEEE}.

\bibitem[\protect\citeauthoryear{Lee \bgroup et al\mbox.\egroup
  }{2014}]{Lee2014}
Lee, C.; Bhardwaj, A.; Di, W.; and andR. Piramuthu, V.~J.
\newblock 2014.
\newblock Region-based discriminative feature pooling for scene text
  recognition.
\newblock IEEE Computer Vision and Pattern Recognition (CVPR).

\bibitem[\protect\citeauthoryear{Lucas \bgroup et al\mbox.\egroup
  }{2003}]{Lucas2003}
Lucas, S.~M.; Panaretos, A.; Sosa, L.; Tang, A.; Wong, S.; and Young, R.
\newblock 2003.
\newblock Icdar 2003 robust reading competitions.
\newblock International Conference on Document Analysis and Recognition
  (ICDAR).

\bibitem[\protect\citeauthoryear{Mishra., Alahari, and
  Jawahar}{2012}]{Mishra2012}
Mishra., A.; Alahari, K.; and Jawahar, C.
\newblock 2012.
\newblock Scene text recognition using higher order language priors.
\newblock British Machine Vision Conference (BMVC).

\bibitem[\protect\citeauthoryear{Neumann and Matas}{2013}]{Neumann2013}
Neumann, L., and Matas, J.
\newblock 2013.
\newblock Scene text localization and recognition with oriented stroke
  detection.
\newblock IEEE International Conference on Computer Vision (ICCV).

\bibitem[\protect\citeauthoryear{Shi \bgroup et al\mbox.\egroup
  }{2013}]{Shi2013}
Shi, C.; Wang, C.; Xiao, B.; Zhang, Y.; Gao, S.; and Zhang, Z.
\newblock 2013.
\newblock Scene text recognition using part-based tree-structured character
  detection.
\newblock IEEE Computer Vision and Pattern Recognition (CVPR).

\bibitem[\protect\citeauthoryear{Su and Lu}{2014}]{Su2014}
Su, B., and Lu, S.
\newblock 2014.
\newblock Accurate scene text recognition based on recurrent neural network.
\newblock Asian Conference on Computer Vision (ICCV).

\bibitem[\protect\citeauthoryear{Sutskever, Vinyals, and
  Le}{2014}]{Sutskever2014}
Sutskever, I.; Vinyals, O.; and Le, Q.~V.
\newblock 2014.
\newblock Sequence to sequence learning with neural networks.
\newblock Neural Information Processing Systems (NIPS).

\bibitem[\protect\citeauthoryear{Visin \bgroup et al\mbox.\egroup
  }{2015}]{Visin2015}
Visin, F.; Kastner, K.; Cho, K.; Matteucci, M.; Courville, A.; and Bengio, Y.
\newblock 2015.
\newblock Renet: A recurrent neural network based alternative to convolutional
  networks.
\newblock {\em arXiv:1505.00393}.

\bibitem[\protect\citeauthoryear{Wang, Babenko, and Belongie}{2011}]{Wang2011}
Wang, K.; Babenko, B.; and Belongie, S.
\newblock 2011.
\newblock End-to-end scene text recognition.
\newblock IEEE International Conference on Computer Vision (ICCV).

\bibitem[\protect\citeauthoryear{Wang \bgroup et al\mbox.\egroup
  }{2012}]{Wang2012}
Wang, T.; Wu, D.; Coates, A.; and Ng, A.~Y.
\newblock 2012.
\newblock End-to-end text recognition with convolutional neural networks.
\newblock IEEE International Conference on Pattern Recognition (ICPR).

\bibitem[\protect\citeauthoryear{Yao \bgroup et al\mbox.\egroup
  }{2014}]{Yao2014}
Yao, C.; Bai, X.; Shi, B.; and Liu, W.
\newblock 2014.
\newblock Strokelets: A learned multi-scale representation for scene text
  recognition.
\newblock IEEE Computer Vision and Pattern Recognition (CVPR).

\bibitem[\protect\citeauthoryear{Yin \bgroup et al\mbox.\egroup
  }{2014}]{Yin2014}
Yin, X.~C.; Yin, X.; Huang, K.; and Hao, H.~W.
\newblock 2014.
\newblock Robust text detection in natural scene images.
\newblock {\em IEEE Trans. Pattern Analysis and Machine Intelligence}
  36:970--983.

\bibitem[\protect\citeauthoryear{Zhang \bgroup et al\mbox.\egroup
  }{2015}]{Zhang2015}
Zhang, Z.; Shen, W.; Yao, C.; and Bai, X.
\newblock 2015.
\newblock Symmetry-based text line detection in natural scenes.
\newblock IEEE Computer Vision and Pattern Recognition (CVPR).

\end{thebibliography}
}

\end{document}